\documentclass[runningheads]{llncs}
\usepackage{graphicx}
\usepackage{multirow}
\usepackage{tikz}
\usepackage{comment}
\usepackage{amsmath,amssymb} % define this before the line numbering.
\usepackage{color}

\usepackage[accsupp]{axessibility}  % Improves PDF readability for those with disabilities.

\begin{document}
% \linenumbers
\pagestyle{headings}
\mainmatter
\def\ECCVSubNumber{004}  % Insert your submission number here

\title{Inspecting Explainability of Transformer Models with Additional Statistical Information} % Replace with your title

\begin{comment}
\titlerunning{ECCV-22 submission ID \ECCVSubNumber} 
\authorrunning{ECCV-22 submission ID \ECCVSubNumber} 
\author{Anonymous RCV submission}
\institute{Paper ID \ECCVSubNumber}
\end{comment}
%******************

%\begin{comment}
\titlerunning{Inspecting Explainability of Transformer Models}
\author{Hoang C. Nguyen \and
Haeil Lee \and
Junmo Kim}
\authorrunning{Nguyen et al.}
\institute{KAIST, Daejeon, Republic of Korea}
%\end{comment}
%******************
\maketitle

\begin{abstract}
Transformer becomes more popular in the vision domain in recent years so there is a need for finding an effective way to interpret the Transformer model by visualizing it. In recent work, Chefer et al. \cite{chefer2021transformer,chefer2021generic} can visualize the Transformer on vision and multi-modal tasks effectively by combining attention layers to show the importance of each image patch. However, when applying to other variants of Transformer such as the Swin Transformer, this method can not focus on the predicted object. Our method, by considering the statistics of tokens in layer normalization layers, shows a great ability to interpret the explainability of Swin Transformer \cite{liu2021swin} and ViT \cite{dosovitskiy2021image}.
\end{abstract}

\section{Introduction}

There is a high demand to understand the decision of AI. The explainability of AI can retain user acceptance and trust \cite{ribeiro2016why} and also helps the developer to guarantee that the model is legal \cite{2017}. Thus explainable AI can help developers to evaluate and debug the model more effectively since they can see which feature the model is based on \cite{ribeiro2016why}.\\
Transformer \cite{vaswani2017attention}, which is the most popular backbone of state-of-the-art models in all NLP tasks also become popular in computer vision in recent years. ViT \cite{dosovitskiy2021image}, Swin Transformer \cite{liu2021swin} can surpass CNN base model while pre-train on a large dataset. Swin Transformer \cite{liu2021swin} also performs well on other downstream tasks such as object detection and image segmentation.\\
Transformer attribution \cite{chefer2021transformer} was developed to create a saliency map for ViT \cite{dosovitskiy2021image}. It outperforms grad cam \cite{DBLP:journals/corr/SelvarajuDVCPB16} which is developed for CNN in visualizing ViT \cite{dosovitskiy2021image} results. Motivated by this, we try this method on Swin Transformer but it can not create a reasonable attention heat map. We observe that this method does not consider the statistical differences of each token in layer normalization layers so we modify it to improve the result. We also conducted some experiments to see whether our method can be applied to ViT \cite{dosovitskiy2021image}.
\section{Related works}

Most methods for understanding the attention base models such as Transformer only use attention scores to show the relationship between each element. These methods, however, ignore all of the other components of self-attention and also other components of the model. \\
Roll out  \cite{abnar2020quantifying} method was introduced on the assumption that the self-attention was stacked linearly so that they can aggregate the attention score from different layers. However, it still ignores other components of Transformer, and also it is not class discriminative that does not know whether the contribution is negative or positive.\\
In Chefer et al. \cite{chefer2021transformer}, they improve the roll out  method by using relevance score (in their later work \cite{chefer2021generic}, they simply use attention matrix) multiple element-wise with the gradient of attention matrix and take the positive value. By using gradient and relevance score, the saliency map is now can get the information from other elements of Transformer. This method outperforms grad cam \cite{2019}, LRP \cite{10.1371/journal.pone.0130140}, and roll out \cite{abnar2020quantifying}  on perturbation and segmentation tasks with ViT \cite{dosovitskiy2021image} in Imagenet.
\section{Method}
\subsection{Swin Transformer explainability}
\begin{figure}
\centering
\includegraphics[width=\textwidth]{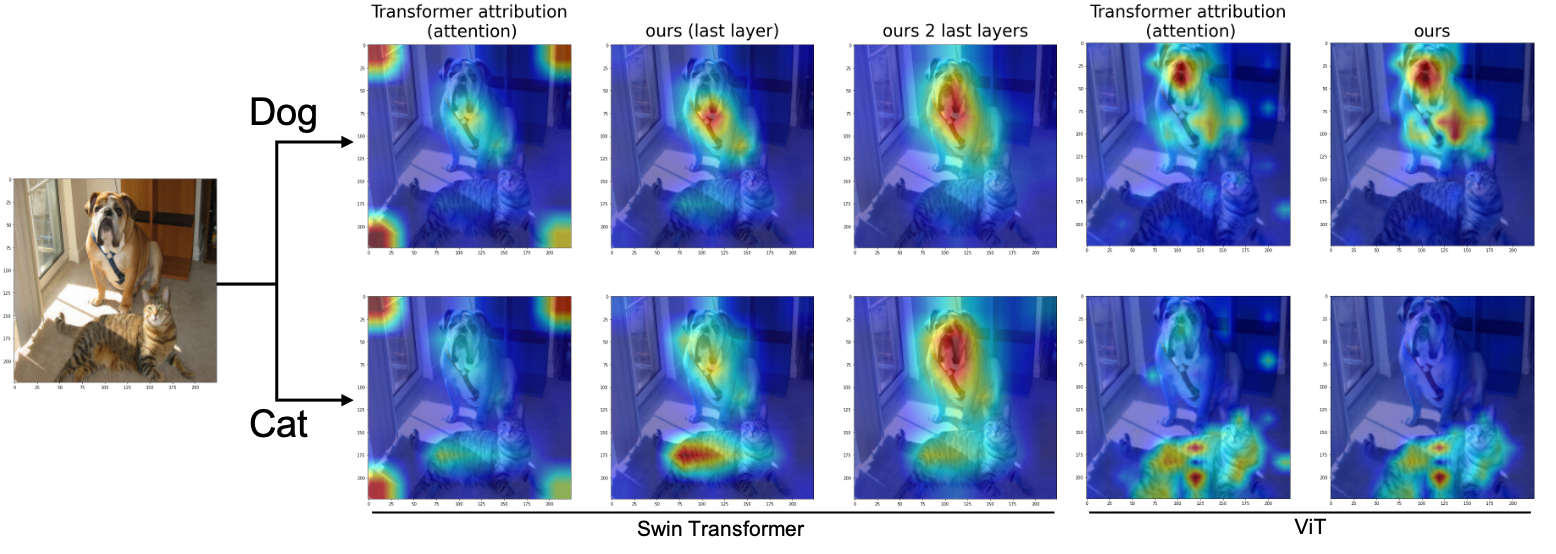}
\caption{Sample visualization results. For Swin Transformer \cite{liu2021swin}, Transformer attribution method shows all the attention comes to the corner. Our method with 2 layers (0 and 1) gets reasonable object location for different targets. On the other hand, ViT \cite{dosovitskiy2021image} visualization, our method reduces the noise on the output heat map.}
\label{fig:figure}
\end{figure}

We first apply the method proposed in Chefer et al. \cite{chefer2021transformer}  to Swin Transformer. However, we only consider the attention matrix instead of computed relevancy since Chefer et al.\cite{chefer2021generic} claim that use of the relevancy matrix can be replaced by an attention matrix. First the relation matrix $R$ is initialized with identity and updated by the following rule.\\
\begin{equation}\label{self1}
    \bar{A}= E_{h}(\nabla A \odot A)^{+}
\end{equation}
\begin{equation}\label{self2}
    R= R+\bar{A} \cdot R  
\end{equation}
In equation \eqref{self1} $A$ is attention matrix and $h$ refer to attention head. The raw attention matrix is multiplied element-wise with its gradient and takes the positive value before taking the average through its multi-head.\\
In Swin Transformer \cite{liu2021swin}, unlike ViT \cite{dosovitskiy2021image}, the number of patches at each layer of Swin Transformer can be different so the size of the attention matrix can be different. Therefore we can not multiply the matrix as proposed in Chefer et al.\cite{chefer2021transformer} to combine the information. To address this problem, we can simply take the average of the rows corresponding to the merging patch as the value for the row corresponding to the merged patch. By doing this, we can calculate the product of the score matrix of two consecutive layers with different feature map sizes. 
First, we just compute the attention interaction for each layer to get the relation matrices: $R^{i}$ using equation \eqref{self1}, \eqref{self2} for $R^{i}$ represent the influence of the input tokens on the output tokens of the layers $i$ through its block. \\
Initialize with $R=R^{j}$ where $j$ is the beginning layers we begin to take the consideration, let $f$ be the taking average function as above we can combine the information by:\\
\begin{equation}\label{swin_update2}
    R= f(R)\cdot R^{i}
\end{equation}
For $i=j+1$ to $d$ for $d$ is the depth of the model.\\
Since the model uses global pooling before the classification head, to make the final heat map, we only need to sum up the value of $L$ at each column. \\
However, when we simply set $j=d$, all the attention comes to $4$ corners of the image as shown in figure \ref{fig:figure}. This phenomenon leads us to the next part of this research.
\subsection{Statistical in Layer normalization}
    In Chefer et al. \cite{chefer2021transformer} and Annar et al. \cite{abnar2020quantifying} only the linear combination of the tokens can be considered and other elements such as FFN or linear transformation  removed since each token is treated the same in these elements. However, Swin Transformer \cite{liu2021swin} and the original Transformer \cite{vaswani2017attention} use layer normalization where each token is divided by its standard deviation (after minus its mean). If the mean effect is removed and follow this intuition, we need to divide the value of each column of the matrix in equation \eqref{self1} by the standard derivation of the corresponding token before adding it with the identity.\\

    \begin{equation}
        \bar{A}= \bar{A} /std(x)
    \end{equation}
    $/$ for column dividing\\
    We also notice that the value of each matrix becomes too small and it can cause unstable computing, we decide to normalize the matrix in \eqref{self1} before adding to the identity so that the sum of the matrix is equal to one.\\
    \begin{equation}
        \bar{A}=\bar{A}/\bar{A}.sum()
    \end{equation}
    
\section{Experiment and result}
We apply Transformer attribution method \cite{chefer2021transformer} and our method to Swin Transformer \cite{liu2021swin} based model for the last 2 layers. The qualitative result is shown in figure \ref{fig:figure}. Also, we run the perturbation test on Imagenet validation dataset \cite{5206848} and the segmentation test on Imagenet segmentation dataset \cite{imagenetseg}, the result is shown in table \ref{table:table}.\\
As we mentioned, the Transformer attribution method \cite{chefer2021transformer} can not localize the object as the patches in the corner has high variance compared with the other region. After dividing the matrix with the standard derivation, the performance improves significantly. Also, when combined with layer 1 of Swin Transformer \cite{liu2021swin}, the result heat map becomes better. \\
We also apply our method to ViT \cite{dosovitskiy2021image} based model. Figure \ref{fig:figure} shows some examples between our method and the original method. Our method removes some noise on the heat map compared to the original method. Also, we run the perturbation test on Imagenet validation dataset \cite{5206848} and the segmentation test on Imagenet segmentation dataset \cite{imagenetseg}. Our method outperforms the original method using only attention matrix and gets the comparable result with Transfomrer attribution method \cite{chefer2021transformer} using complex relevance matrix.\\

\setlength{\tabcolsep}{4pt}
\begin{table}
\begin{center}
\caption{Segmentation and perturbation test result on Imagenet validation dataset \cite{imagenetseg} for Swin Transformer \cite{liu2021swin} (upper rows) and ViT \cite{dosovitskiy2021image} (lower rows).} 
\label{table:table}
% \usepackage{multirow}
\begin{tabular}{c|cccc|cccc}
\hline
\multirow{2}{*}{Method}  & \multicolumn{4}{c|}{Segmentation}                                            & \multicolumn{4}{c}{Perturbation}                                                                                                                                                                                     \\
                         & mIoU  & mAP   & \begin{tabular}[c]{@{}c@{}}Pixel \\ Acc\end{tabular} & mF1   & \begin{tabular}[c]{@{}c@{}}Top \\ Neg\end{tabular} & \begin{tabular}[c]{@{}c@{}}Top\\ Pos\end{tabular} & \begin{tabular}[c]{@{}c@{}}Target\\ Neg\end{tabular} & \begin{tabular}[c]{@{}c@{}}Target\\ Pos\end{tabular} \\ \hline
Attn Layer Norm(ours)    & 49.38 & 79.28 & 69.75                                                & 34.82 & 62.11                                              & 37.83                                             & 62.51                                                & 37.50                                                \\
Attn                     & 39.06 & 70.55 & 62.63                                                & 22.80 & 57.11                                              & 43.17                                             & 57.64                                                & 42.66                                                \\
Attn Layer Norm(Layer1) & 54.19 & 83.63 & 72.78                                                & 38.23 & 66.87                                              & 35.98                                             & 67.38                                                & 35.59                                                \\ \hline
Transformer Attribution  & 61.98 & 86.04 & 79.72                                                & 40.18 & 54.16                                              & 17.03                                             & 55.04                                                & 16.04                                                \\
Attn Layer Norm(ours)    & 62.95 & 86.81 & 79.98                                                & 42.30 & 55.40                                              & 17.10                                             & 56.42                                                & 16.46                                                \\
Attn                     & 58.34 & 85.28 & 76.30                                                & 41.85 & 54.61                                              & 17.32                                             & 55.67                                                & 16.72                                                \\ \hline
\end{tabular}
\end{center}
\end{table}
\setlength{\tabcolsep}{1.4pt}

% \begin{figure}[t]
% \includegraphics[width=\textwidth]{swin visualization.png}
% \caption{Sample result for Swin visualization. For Transformer attribution method all the attention comes to the corner. Our method with 2 layers (0 and 1) get compatible result with grad cam on model norm layers}
% \label{fig:figure1}
% \end{figure}

% \begin{figure}[t]
% \includegraphics[width=\textwidth]{vit_visualization.png}
% \caption{Sample result for VIT visualization. Our method reduces the noise on the output heat map. Our method improves roll out method also }
% \label{fig:figure2}
% \end{figure}

\section{Conclusion}
Transformer attribution \cite{chefer2021transformer,chefer2021generic} provides a very intuitive and effective way to visualize Transformer. However, it cannot work well on other Transformer variants such as Swin Transformer \cite{liu2021swin}. Our proposed method can improve visualizing explainability of Transformer model on Swin Transformer \cite{liu2021swin} and ViT \cite{dosovitskiy2021image} itself by considering the statistical differences of each token in layer normalization layers.

\clearpage
\bibliographystyle{splncs04}
\bibliography{egbib}

\begin{thebibliography}{10}
\providecommand{\url}[1]{\texttt{#1}}
\providecommand{\urlprefix}{URL }
\providecommand{\doi}[1]{https://doi.org/#1}

\bibitem{abnar2020quantifying}
Abnar, S., Zuidema, W.: Quantifying attention flow in transformers (2020)

\bibitem{10.1371/journal.pone.0130140}
Bach, S., Binder, A., Montavon, G., Klauschen, F., Müller, K.R., Samek, W.: On pixel-wise explanations for non-linear classifier decisions by layer-wise relevance propagation. PLOS ONE  \textbf{10}(7),  1--46 (07 2015). \doi{10.1371/journal.pone.0130140}, \url{https://doi.org/10.1371/journal.pone.0130140}

\bibitem{chefer2021generic}
Chefer, H., Gur, S., Wolf, L.: Generic attention-model explainability for interpreting bi-modal and encoder-decoder transformers (2021)

\bibitem{chefer2021transformer}
Chefer, H., Gur, S., Wolf, L.: Transformer interpretability beyond attention visualization (2021)

\bibitem{5206848}
Deng, J., Dong, W., Socher, R., Li, L.J., Li, K., Fei-Fei, L.: Imagenet: A large-scale hierarchical image database. In: 2009 IEEE Conference on Computer Vision and Pattern Recognition. pp. 248--255 (2009). \doi{10.1109/CVPR.2009.5206848}

\bibitem{dosovitskiy2021image}
Dosovitskiy, A., Beyer, L., Kolesnikov, A., Weissenborn, D., Zhai, X., Unterthiner, T., Dehghani, M., Minderer, M., Heigold, G., Gelly, S., Uszkoreit, J., Houlsby, N.: An image is worth 16x16 words: Transformers for image recognition at scale (2021)

\bibitem{2017}
Goodman, B., Flaxman, S.: European union regulations on algorithmic decision-making and a “right to explanation”. AI Magazine  \textbf{38}(3),  50–57 (Oct 2017). \doi{10.1609/aimag.v38i3.2741}, \url{http://dx.doi.org/10.1609/aimag.v38i3.2741}

\bibitem{imagenetseg}
Guillaumin, M., K\"{u}ttel, D., Ferrari, V.: Imagenet auto-annotation with segmentation propagation. Int. J. Comput. Vision  \textbf{110}(3),  328–348 (dec 2014). \doi{10.1007/s11263-014-0713-9}, \url{https://doi.org/10.1007/s11263-014-0713-9}

\bibitem{liu2021swin}
Liu, Z., Lin, Y., Cao, Y., Hu, H., Wei, Y., Zhang, Z., Lin, S., Guo, B.: Swin transformer: Hierarchical vision transformer using shifted windows (2021)

\bibitem{ribeiro2016why}
Ribeiro, M.T., Singh, S., Guestrin, C.: "why should i trust you?": Explaining the predictions of any classifier (2016)

\bibitem{2019}
Selvaraju, R.R., Cogswell, M., Das, A., Vedantam, R., Parikh, D., Batra, D.: Grad-cam: Visual explanations from deep networks via gradient-based localization. International Journal of Computer Vision  \textbf{128}(2),  336–359 (Oct 2019). \doi{10.1007/s11263-019-01228-7}, \url{http://dx.doi.org/10.1007/s11263-019-01228-7}

\bibitem{DBLP:journals/corr/SelvarajuDVCPB16}
Selvaraju, R.R., Das, A., Vedantam, R., Cogswell, M., Parikh, D., Batra, D.: Grad-cam: Why did you say that? visual explanations from deep networks via gradient-based localization. CoRR  \textbf{abs/1610.02391} (2016), \url{http://arxiv.org/abs/1610.02391}

\bibitem{vaswani2017attention}
Vaswani, A., Shazeer, N., Parmar, N., Uszkoreit, J., Jones, L., Gomez, A.N., Kaiser, L., Polosukhin, I.: Attention is all you need (2017)

\end{thebibliography}
\end{document}